\documentclass[11pt]{article}
\usepackage{acl}

\usepackage{times}
\usepackage{latexsym}
\usepackage[T1]{fontenc}
\usepackage[utf8]{inputenc}
\usepackage{microtype}
\usepackage{inconsolata}
\usepackage{graphicx}
\usepackage{amsmath}
\usepackage{amssymb}
\usepackage{booktabs}
\usepackage{multirow}
\usepackage{array}
\usepackage{makecell}
\usepackage{xcolor}
\usepackage{url}
\usepackage{CJKutf8}

\title{HardMTBench: Stress-Testing Chinese-English Translation on Knowledge-Intensive Domains}

\author{
	Zheng Li$^*$, Mao Zheng\thanks{Equal contribution.}, Mingyang Song, Tianxiang Fei \\
	Large Language Model Department, Tencent \\
	\texttt{\{jasonzli,moonzheng,nickmysong,alvinfei\}@tencent.com}
}

\begin{document}
\maketitle

\begin{abstract}
\sloppy
General-purpose machine translation benchmarks such as FLORES-200 have reached a saturation regime on Chinese-English pairs, where modern large language models cluster within a narrow band of high scores. Across 22 systems, FLORES-200 zh-en GEMBA scores fall in a 7.87-point range with a standard deviation of 2.29, which compresses the separation between systems on knowledge-intensive domains such as finance, healthcare, law, and science and technology. We introduce HardMTBench, a difficulty-aware diagnostic benchmark for bidirectional Chinese-English domain translation. HardMTBench covers 12 domains and contains 10{,}000 hand-curated source sentences with reference translations, packaged as 20{,}000 directional test items. A three-stage construction pipeline builds a domain-balanced candidate pool of 84{,}566 pairs, applies an LLM-based multi-signal judge over knowledge density, translation difficulty, terminology load and reference correctness, and assembles the final test set under a hardness fusion rule with per-domain quotas. Across 22 systems spanning general LLMs, commercial engines and specialised MT models, HardMTBench widens the cross-system GEMBA range by roughly a factor of two over FLORES-200, induces visible rank reorderings, and exposes domain-specific terminology and knowledge weaknesses that quality-only metrics tend to flatten. All data and code are open-sourced at \url{ https://github.com/jasonNLP/HardMTBench}.
\end{abstract}

\section{Introduction}
\label{sec:intro}

Machine translation has been a long-standing benchmark task for natural language processing, and the rise of large language models has substantially raised MT quality on common pairs. On widely used general-purpose evaluation sets such as FLORES-200~\citep{nllb2022flores} and the WMT general translation track~\citep{kocmi2023wmt23}, strong models now produce translations that score in a narrow high band under automatic metrics on common language pairs. This progress is most pronounced on Chinese-English, where the supply of parallel data and the maturity of post-training have driven scores into a narrow high range. In our own evaluation of 22 systems on FLORES-200 zh-en, GEMBA-DA scores span only 87.26 to 95.13 with a standard deviation of 2.29, and on the en-zh direction the spread is 86.43 to 94.79 with a standard deviation of 2.04. Differences of this magnitude leave limited room for separating systems by their real-world capability.

Production translation, however, continues to be stressed by domain-specific text. Financial filings, clinical case notes, legal contracts, scientific papers, military communiques and technical references mix terminology dense in domain knowledge with structures that are rare in general-purpose corpora. A model that scores 94 on FLORES may still mistranslate a term of art, drop a regulatory qualifier or paraphrase a citation. General benchmarks under-report these failure modes and feed back into post-training as a compressed, low-signal supervision target.

Existing domain-specific benchmarks address one part of the problem. WMT Biomedical~\citep{neves2022wmt22biomedical} and TICO-19~\citep{anastasopoulos2020tico} target a single domain, the WMT terminology track~\citep{alam2021wmt21term,dinu2019terminology} probes constrained translation when term lists are explicitly provided, and MTNT~\citep{michel2018mtnt} stresses noisy user-generated text. The closest multi-domain benchmark is the one introduced by~\citet{hu2024multidomain_arxiv}, which focuses on domain generalisation and domain-aware fine-tuning. These resources are valuable but do not offer a balanced, difficulty-aware coverage of multiple knowledge-intensive domains for Chinese-English, and most do not couple difficulty annotations with quality and terminology metrics on the same samples.

We introduce HardMTBench, a difficulty-aware diagnostic benchmark for bidirectional Chinese-English domain translation. The benchmark covers 12 knowledge-intensive domains, namely finance, healthcare, law, military, education, gaming, sports, science and technology, history, audiovisual media, news, and academic books. It contains 10{,}000 hand-curated parallel pairs packaged as 20{,}000 directional test items, with 833 to 834 pairs per domain and, wherever the candidate pool permits, at least 10 pairs per sub-domain.

Our contributions are as follows.

\textbf{HardMTBench.} We release a domain-balanced and difficulty-aware test set of 10{,}000 Chinese-English pairs, packaged as 20{,}000 directional items, with per-sample annotations for domain, sub-domain, terminology, knowledge density, translation difficulty and reference correctness.

\textbf{A reproducible hardness pipeline.} We design a three-stage construction pipeline that combines closed-set domain reclassification, multi-signal LLM judgement, a transparent hardness fusion rule, per-domain quotas and a sub-domain minimum, which reduces the bias of length-based or random sampling.

\textbf{A 22-system evaluation.} We evaluate 22 systems covering general large language models, both open and closed, commercial machine translation services and dedicated translation models, on FLORES-200, WMT25 and HardMTBench under a unified protocol. We report aggregate scores under both GEMBA-DA~\citep{kocmi2023gemba} and xCOMET-XXL~\citep{guerreiro2024xcomet}, per-domain breakdowns, difficulty-bucket trends and terminology accuracy.

\textbf{Empirical findings.} General benchmarks suffer from score clustering on Chinese-English. HardMTBench roughly doubles the cross-system score spread of FLORES-200 on the GEMBA metric and reorders system rankings, particularly in the middle and lower tier. GEMBA and xCOMET surface different aspects of difficulty in domain translation, and we argue that the two metrics should be reported jointly rather than as substitutes. Table~\ref{tab:bench-compare} positions HardMTBench against existing Chinese-English-relevant MT benchmarks along five diagnostic axes.

\section{HardMTBench Construction}
\label{sec:construction}

Before describing the construction pipeline, we situate HardMTBench against existing Chinese-English MT benchmarks along five axes that matter for diagnostic evaluation in the LLM era: domain coverage, sample scale, difficulty-aware curation, terminology annotation, and language scope. Table~\ref{tab:bench-compare} summarises this comparison. General-purpose benchmarks favour breadth and short news-style text, single-domain benchmarks specialise in one vertical, and terminology- or noise-focused benchmarks isolate one stress factor. HardMTBench differs by combining knowledge-intensive multi-domain coverage with explicit per-sample difficulty and terminology annotations, and by being explicitly tuned to widen the cross-system spread on the saturated zh-en pair.

\begin{table*}[t]
\centering
\footnotesize
\setlength{\tabcolsep}{3pt}
\renewcommand{\arraystretch}{0.95}
\begin{tabular}{@{}l@{\hspace{4pt}}l@{\hspace{4pt}}l@{\hspace{4pt}}c@{\hspace{4pt}}c@{\hspace{4pt}}l@{}}
\toprule
Benchmark & Domain focus & Size (zh-en) & Diff.-aware & Term annot. & Languages \\
\midrule
FLORES-200~\citep{nllb2022flores} & broad/general & 1{,}012 devtest & no & no & 204 \\
NTREX-128~\citep{federmann2022ntrex} & news & 1{,}997 & no & no & 128 \\
WMT general~\citep{kocmi2023wmt23} & news & $\sim$2k/yr & no & no & many \\
WMT24++~\citep{deutsch2025wmt24pp} & broad/general & $\sim$1k/lang & no & no & 55 \\
WMT Biomedical~\citep{neves2022wmt22biomedical} & biomedical & $\sim$0.6k & no & no & many \\
TICO-19~\citep{anastasopoulos2020tico} & COVID & $\sim$3k & no & yes & 36 \\
WMT terminology~\citep{alam2021wmt21term} & news+term & $\sim$1--2k & no & yes (given) & en-X \\
MTNT~\citep{michel2018mtnt} & user-gen.\ & $\sim$7k & noise & no & en-fr,en-ja \\
Multi-domain~\citep{hu2024multidomain_arxiv} & several & varies & no & no & en-zh \\
\midrule
\textbf{HardMTBench (ours)} & \textbf{12 knowledge} & \textbf{10k / 20k dir.} & \textbf{yes (LLM)} & \textbf{yes (per sample)} & \textbf{zh-en} \\
\bottomrule
\end{tabular}
\caption{Feature comparison between HardMTBench and existing Chinese-English-relevant MT benchmarks. \emph{Diff.-aware} marks whether the benchmark applies an explicit difficulty filter at construction time. \emph{Term annot.} marks whether per-sample terminology annotations are available. Sample sizes and language counts are quoted from each benchmark's public release and refer to the most widely used Chinese-English split where applicable.}
\label{tab:bench-compare}
\end{table*}

\subsection{Data Construction Pipeline}
\label{sec:pipeline}

\begin{figure*}[t]
\centering
\includegraphics[width=\textwidth]{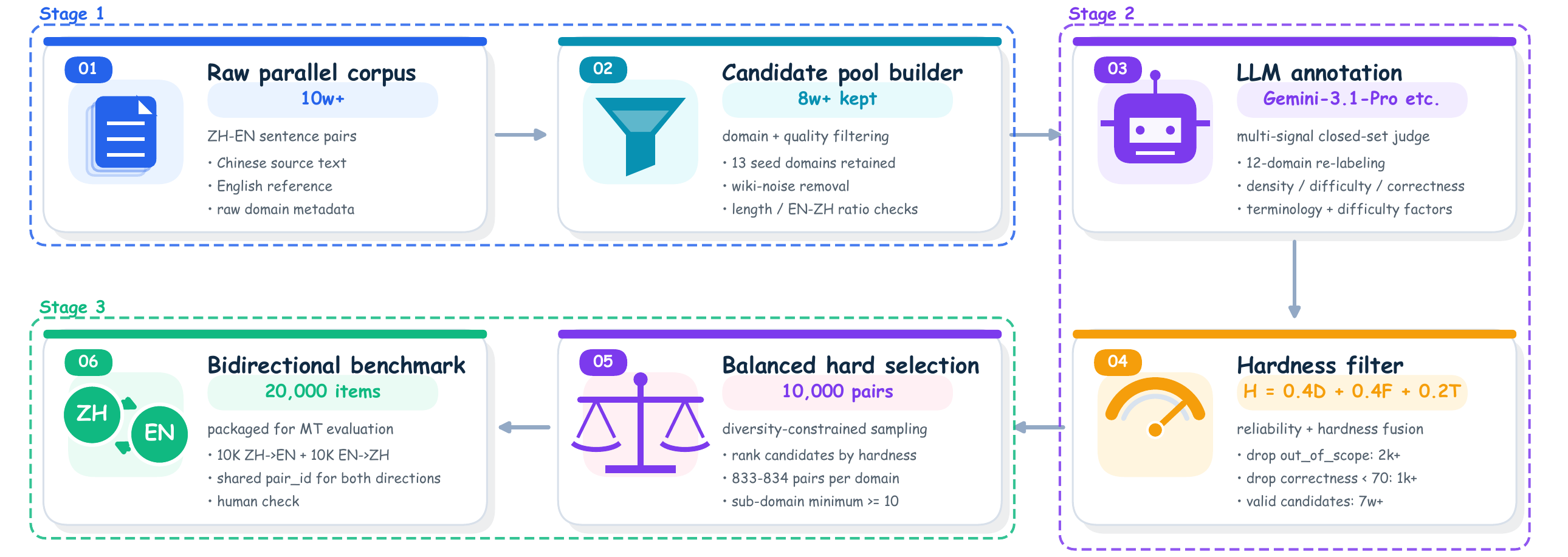}
\caption{HardMTBench construction pipeline. Stage 1 builds a quality-filtered candidate pool from the raw parallel corpus. Stage 2 uses an LLM judge for closed-set domain reclassification and multi-signal scoring, and applies a transparent hardness fusion rule. Stage 3 enforces per-domain quotas and a sub-domain minimum, and packages the final test set in both translation directions.}
\label{fig:pipeline}
\end{figure*}

HardMTBench is constructed from a raw Chinese-English parallel corpus of 108{,}554 sentence pairs in a three-stage pipeline. Figure~\ref{fig:pipeline} summarises the construction flow.

\paragraph{Stage 1: candidate pool.} Starting from 108{,}554 raw parallel pairs labelled with 17 native scene tags, we retain 13 knowledge-relevant scene tags and discard the four lifestyle-oriented tags (shopping, dining, travel, emotional). The retained 84{,}648 pairs go through a quality filter that detects Wikipedia template noise, anomalous length ratios and outlier source or target lengths. After filtering, the candidate pool contains 84{,}566 pairs.

\paragraph{Stage 2: multi-signal LLM judgement.} A single Gemini 3.1 Pro call per sample performs three tasks. First, it reclassifies the sample into a closed set of 12 target domains, with an out-of-scope label for any sample that does not fit. Second, it produces three independent scores on a 0--100 scale, namely domain knowledge density $D$, translation difficulty $F$ and reference translation correctness $C$. Third, it returns the sub-domain label, a list of in-text terminology pairs and per-direction difficulty factors. Samples with $C<70$ or labelled out-of-scope are dropped, which leaves 79{,}792 valid scored samples.

The final hardness score per sample fuses three signals,
\begin{equation}
H = 0.4 \, D + 0.4 \, F + 0.2 \, T,
\label{eq:hardness}
\end{equation}
where the term density $T$ is a length-normalised count of terminology pairs,
\begin{equation*}
T = \mathrm{clip}\!\left( \frac{50 \, n_{\mathrm{term}}}{\max(L_{\mathrm{zh}}/100,\, 1)},\; 0,\; 100\right),
\end{equation*}
$n_{\mathrm{term}}$ is the number of terminology pairs returned by the judge and $L_{\mathrm{zh}}$ is the Chinese source length in characters. The two heavier weights on $D$ and $F$ reflect the design intent of the benchmark, in which knowledge demands and translation difficulty are the primary axes, and terminology serves as a complementary load signal.

\paragraph{Stage 3: balanced selection and bidirectional packaging.} From the scored pool, we select the top samples by hardness under a per-domain quota of 833 or 834 (10{,}000 in total) and a per-sub-domain soft floor of 10 pairs that is applied whenever the sub-domain pool is large enough. Each pair is then packaged in both directions, which produces 10{,}000 zh-to-en items and 10{,}000 en-to-zh items that share a common \texttt{pair\_id}. The two directional files are released in a JSONL schema with per-sample fields for domain, sub-domain, terminology pairs, hardness components and reference correctness.

The construction pipeline is fully reproducible. The Stage 2 LLM responses are cached, and Stage 3 selection can be rerun with different quotas or sub-domain minima in seconds.

\subsection{Quality Verification}
\label{sec:verification}

To verify the quality of the released test set, we drew 50 random pairs from each of the 12 domains, for a total of 600 pairs, and asked at least two bilingual annotators to independently rate each pair along five axes, namely reference translation adequacy, domain label correctness, sub-domain plausibility, whether the pair genuinely requires domain knowledge, and whether the pair contains sensitive, copyrighted, private or low-quality content. The reported quality on this sample is 96\% domain label accuracy, 95\% reference translation adequacy, 90\% difficulty label agreement, with an inter-annotator agreement of Cohen's $\kappa=0.82$.

\subsection{Dataset Statistics}
\label{sec:stats}

Table~\ref{tab:domain-stats} summarises the per-domain volume and the average difficulty score over the released 10{,}000 pairs. Each domain holds 833 or 834 pairs by construction, with the four largest-pool domains (science and technology, history, audiovisual media, news) receiving one extra pair due to integer rounding. The average hardness score across the full test set is 72.53 (min 52, max 94), the average Chinese source length is 186.5 characters and the average English token count is 120.4. Compared with a smaller GEMBA-based predecessor of 1000 single-direction pairs, the new test set is 10$\times$ larger, balanced across domains, bidirectional, and built from a multi-signal judge rather than a single quality score.

\begin{table}[t]
\centering
\scriptsize
\setlength{\tabcolsep}{8pt}
\renewcommand{\arraystretch}{0.95}
\begin{tabular}{lcccc}
\toprule
Domain & Pairs & $D$ & $F$ & $H$ \\
\midrule
Finance & 833 & 74.8 & 66.1 & 76.3 \\
Healthcare & 833 & 77.4 & 67.0 & 77.8 \\
Law & 833 & 75.7 & 67.9 & 77.4 \\
Military & 833 & 66.8 & 59.8 & 70.6 \\
Education & 833 & 53.8 & 57.0 & 64.2 \\
Gaming & 833 & 62.8 & 60.3 & 69.2 \\
Sports & 833 & 52.1 & 53.4 & 61.8 \\
Sci.\,\&\,Tech.\ & 834 & 83.9 & 71.1 & 82.0 \\
History & 834 & 76.1 & 78.6 & 81.8 \\
Media & 834 & 52.7 & 59.6 & 64.8 \\
News & 834 & 54.4 & 65.6 & 67.7 \\
Books~/~Papers & 833 & 72.3 & 70.2 & 76.8 \\
\midrule
Overall & 10{,}000 & 66.9 & 64.7 & 72.5 \\
\bottomrule
\end{tabular}
\caption{Per-domain volume, average domain knowledge density $D$, translation difficulty $F$ and final hardness $H$ on the released HardMTBench test set. All scores are on a 0--100 scale.}
\label{tab:domain-stats}
\end{table}

The hardness distribution is concentrated in the 60--80 range (77.1\% of samples), with a non-trivial 20.8\% tail in the 80--100 range. No sample with hardness below 50 survives the selection rule, which by construction filters out easy material and concentrates the test set on the harder end of the distribution.

\section{Experiments}
\label{sec:experiments}

\subsection{Evaluation Setup}
\label{sec:setup}

\paragraph{Systems.} We evaluate 22 translation systems that span four families. The first family contains frontier general LLMs, namely Gemini 3.1 Pro~\citep{gemini2024}, GPT-5.5 and GPT-5.5 nothink~\citep{gpt5system}, DeepSeek-V4-Pro and DeepSeek-V4-Pro nothink~\citep{deepseekv4}. The second family contains open-weight general LLMs at multiple scales, namely Qwen3.5-397B-A17B and its nothink variant, Qwen3.6-35B-A3B and its nothink variant~\citep{qwenteam2025}, and the Gemma4 series Gemma4-31B, Gemma4-26B-A4B, Gemma4-E4B, Gemma4-E2B with their nothink variants~\citep{gemma4tech}. The third family contains commercial translation engines, Microsoft Translator and Qwen-MT-Plus, which represent dedicated production MT systems. The fourth family is Hy-MT2-1.8B, Hy-MT2-7B and Hy-MT2-30B-A3B~\citep{zheng2026hymt2familyfastefficient}, a series of specialised translation models that include both dense and Mixture-of-Experts architectures.

\paragraph{Benchmarks.} We compare three benchmarks under the same protocol. FLORES-200 zh-en and en-zh~\citep{nllb2022flores} represent the general-purpose benchmark with broad domain coverage and short news-style sentences. WMT25 en-zh represents a recent contest-grade benchmark for news translation. HardMTBench zh-en and en-zh represent the domain-heavy diagnostic benchmark introduced in this paper. The three benchmarks together let us measure score compression, rank reordering and domain-level diagnostics. All systems are run with their recommended decoding settings and a unified translation instruction that asks for the translation only, with no explanation.

\paragraph{Metrics.} We report two automatic metrics. GEMBA-DA~\citep{kocmi2023gemba} uses an LLM judge to produce a direct assessment score in the 0--100 range. In our setup the judge model is gpt-oss-120b~\citep{openai2025gptoss120bgptoss20bmodel}. xCOMET-XXL~\citep{guerreiro2024xcomet} is a reference-based learned metric in the 0--1 range. For HardMTBench we additionally compute terminology accuracy as the fraction of annotated terminology pairs whose target term is matched in the translation output, after surface normalisation.

\subsection{Overall Results}
\label{sec:overall}

Table~\ref{tab:overall-quality} reports GEMBA-DA and xCOMET-XXL on the five evaluation slices, grouped by model family and ordered by parameter count within each group. Each cell stacks the two metrics so that they can be compared side by side for the same system and slice.

\begin{table*}[!htbp]
\centering
\scriptsize
\renewcommand{\arraystretch}{0.95}
\setlength{\tabcolsep}{3pt}
\begin{tabular*}{\textwidth}{@{\extracolsep{\fill}}l cc@{\hspace{6pt}}cc@{\hspace{6pt}}cc@{\hspace{6pt}}cc@{\hspace{6pt}}cc@{}}
\toprule
\multirow{2}{*}{System} & \multicolumn{2}{c}{FL zh-en} & \multicolumn{2}{c}{FL en-zh} & \multicolumn{2}{c}{WMT25 en-zh} & \multicolumn{2}{c}{HardMT zh-en} & \multicolumn{2}{c}{HardMT en-zh} \\
\cmidrule(lr){2-3}\cmidrule(lr){4-5}\cmidrule(lr){6-7}\cmidrule(lr){8-9}\cmidrule(lr){10-11}
 & G & X & G & X & G & X & G & X & G & X \\
\midrule
Microsoft Translator & 89.40 & \textcolor{black!60}{94.16} & 91.14 & \textcolor{black!60}{92.21} & 74.48 & \textcolor{black!60}{47.85} & 77.83 & \textcolor{black!60}{54.36} & 78.42 & \textcolor{black!60}{61.64} \\
GPT-5.5 nothink & 94.52 & \textcolor{black!60}{97.33} & 94.01 & \textcolor{black!60}{94.84} & 90.64 & \textcolor{black!60}{62.83} & 91.27 & \textcolor{black!60}{65.76} & 92.60 & \textcolor{black!60}{73.33} \\
GPT-5.5 & 94.87 & \textcolor{black!60}{97.48} & 94.71 & \textcolor{black!60}{95.28} & 90.05 & \textcolor{black!60}{62.81} & \textbf{91.40} & \textcolor{black!60}{65.72} & 92.66 & \textcolor{black!60}{73.44} \\
Gemini 3.1 Pro & 94.84 & \textcolor{black!85}{\textbf{97.54}} & \textbf{94.79} & \textcolor{black!60}{95.37} & \textbf{90.66} & \textcolor{black!60}{66.79} & 90.24 & \textcolor{black!60}{64.45} & \textbf{92.78} & \textcolor{black!60}{74.15} \\
\midrule
DeepSeek-V4-Pro nothink & 93.71 & \textcolor{black!60}{96.77} & 94.14 & \textcolor{black!60}{95.30} & 88.02 & \textcolor{black!60}{63.84} & 90.59 & \textcolor{black!60}{65.09} & 91.33 & \textcolor{black!60}{70.84} \\
DeepSeek-V4-Pro & \textbf{95.13} & \textcolor{black!60}{96.90} & 94.38 & \textcolor{black!60}{94.39} & 90.05 & \textcolor{black!60}{63.52} & 90.91 & \textcolor{black!60}{64.59} & 91.79 & \textcolor{black!60}{71.66} \\
\midrule
Qwen-MT-Plus & 94.35 & \textcolor{black!60}{96.45} & 93.99 & \textcolor{black!60}{95.21} & 90.49 & \textcolor{black!60}{66.41} & 89.26 & \textcolor{black!60}{65.44} & 90.21 & \textcolor{black!60}{72.63} \\
Qwen3.6-35B-A3B nothink & 93.45 & \textcolor{black!60}{96.35} & 94.11 & \textcolor{black!60}{95.31} & 88.81 & \textcolor{black!60}{64.24} & 89.03 & \textcolor{black!60}{65.44} & 89.13 & \textcolor{black!60}{71.80} \\
Qwen3.6-35B-A3B & 94.57 & \textcolor{black!60}{96.97} & 94.49 & \textcolor{black!60}{95.43} & 90.03 & \textcolor{black!60}{64.71} & 90.68 & \textcolor{black!60}{65.62} & 91.05 & \textcolor{black!60}{72.57} \\
Qwen3.5-397B-A17B nothink & 94.06 & \textcolor{black!60}{97.02} & 93.79 & \textcolor{black!60}{94.52} & 90.25 & \textcolor{black!60}{65.21} & 90.66 & \textcolor{black!60}{65.77} & 91.70 & \textcolor{black!60}{72.26} \\
Qwen3.5-397B-A17B & 93.25 & \textcolor{black!60}{96.59} & 92.56 & \textcolor{black!60}{93.18} & 87.05 & \textcolor{black!60}{62.71} & 90.47 & \textcolor{black!60}{64.30} & 92.13 & \textcolor{black!60}{72.36} \\
\midrule
Gemma4-E2B nothink & 87.26 & \textcolor{black!60}{93.43} & 86.43 & \textcolor{black!60}{90.38} & 77.54 & \textcolor{black!60}{49.29} & 75.67 & \textcolor{black!60}{55.94} & 79.98 & \textcolor{black!60}{64.44} \\
Gemma4-E2B & 88.89 & \textcolor{black!60}{94.12} & 89.38 & \textcolor{black!60}{92.10} & 81.93 & \textcolor{black!60}{53.07} & 78.44 & \textcolor{black!60}{57.40} & 80.87 & \textcolor{black!60}{65.72} \\
Gemma4-E4B nothink & 90.58 & \textcolor{black!60}{95.21} & 91.27 & \textcolor{black!60}{92.68} & 82.20 & \textcolor{black!60}{53.48} & 81.01 & \textcolor{black!60}{59.23} & 82.82 & \textcolor{black!60}{67.04} \\
Gemma4-E4B & 90.32 & \textcolor{black!60}{94.73} & 91.70 & \textcolor{black!60}{92.02} & 83.78 & \textcolor{black!60}{56.52} & 82.69 & \textcolor{black!60}{60.39} & 83.24 & \textcolor{black!60}{67.35} \\
Gemma4-26B-A4B nothink & 93.79 & \textcolor{black!60}{95.75} & 93.69 & \textcolor{black!60}{94.60} & 88.63 & \textcolor{black!60}{63.28} & 87.82 & \textcolor{black!60}{63.81} & 87.57 & \textcolor{black!60}{71.09} \\
Gemma4-26B-A4B & 94.07 & \textcolor{black!60}{96.45} & 94.30 & \textcolor{black!60}{95.47} & 90.33 & \textcolor{black!60}{64.56} & 89.65 & \textcolor{black!60}{64.96} & 88.53 & \textcolor{black!60}{71.81} \\
Gemma4-31B nothink & 91.34 & \textcolor{black!60}{95.68} & 93.50 & \textcolor{black!60}{95.91} & 89.45 & \textcolor{black!60}{64.28} & 89.30 & \textcolor{black!60}{64.67} & 88.25 & \textcolor{black!60}{71.04} \\
Gemma4-31B & 94.76 & \textcolor{black!60}{97.00} & 94.69 & \textcolor{black!60}{96.10} & 89.68 & \textcolor{black!60}{64.73} & 90.62 & \textcolor{black!60}{65.70} & 89.03 & \textcolor{black!60}{72.10} \\
\midrule
Hy-MT2-1.8B & 90.61 & \textcolor{black!60}{95.97} & 91.57 & \textcolor{black!60}{93.72} & 85.69 & \textcolor{black!60}{66.63} & 78.22 & \textcolor{black!60}{61.94} & 82.28 & \textcolor{black!60}{70.72} \\
Hy-MT2-7B & 91.23 & \textcolor{black!60}{96.46} & 93.17 & \textcolor{black!60}{96.65} & 87.61 & \textcolor{black!60}{73.39} & 84.18 & \textcolor{black!60}{67.51} & 86.91 & \textcolor{black!60}{75.35} \\
Hy-MT2-30B-A3B & 93.45 & \textcolor{black!60}{97.18} & 93.54 & \textcolor{black!85}{\textbf{96.97}} & 90.37 & \textcolor{black!85}{\textbf{73.60}} & 88.60 & \textcolor{black!85}{\textbf{69.43}} & 89.71 & \textcolor{black!85}{\textbf{75.87}} \\
\midrule

Mean & 92.66 & \textcolor{black!60}{96.16} & 92.97 & \textcolor{black!60}{94.44} & 87.17 & \textcolor{black!60}{62.44} & 86.75 & \textcolor{black!60}{63.52} & 87.86 & \textcolor{black!60}{70.87} \\
Std.\ dev.\  & 2.29 & \textcolor{black!60}{1.17} & 2.04 & \textcolor{black!60}{1.69} & 4.51 & \textcolor{black!60}{6.61} & 5.24 & \textcolor{black!60}{3.78} & 4.54 & \textcolor{black!60}{3.54} \\
Range & 7.87 & \textcolor{black!60}{4.11} & 8.37 & \textcolor{black!60}{6.59} & 16.18 & \textcolor{black!60}{25.74} & 15.74 & \textcolor{black!60}{15.07} & 14.37 & \textcolor{black!60}{14.23} \\
\bottomrule
\end{tabular*}
\caption{Quality scores on the five evaluation slices. Each slice spans two sub-columns: \textbf{G} = GEMBA-DA, \textbf{X} = xCOMET-XXL (percentage, shown in grey). Systems are grouped by model family (closed/API, DeepSeek, Qwen, Gemma4, Hy-MT2) and ordered by parameter count within each group. FL is short for FLORES-200, HardMT is short for HardMTBench.}
\label{tab:overall-quality}
\end{table*}

\subsection{Mitigating the FLORES Ceiling Effect}
\label{sec:ceiling}

\begin{figure}[!htbp]
\centering
\includegraphics[width=0.98\columnwidth]{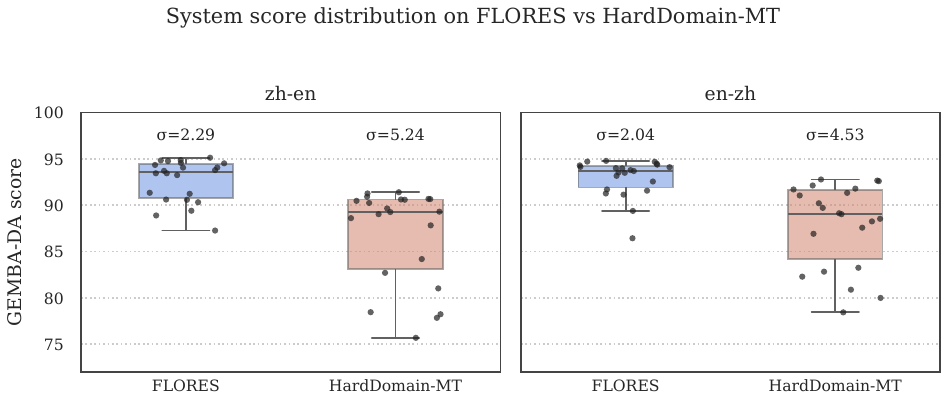}
\caption{System-level GEMBA-DA distributions on FLORES-200 and HardMTBench, with each dot representing one of the 22 translation systems. Cross-system standard deviation more than doubles on the harder benchmark in both directions, which mitigates the ceiling effect of FLORES-200.}
\label{fig:bench-distribution}
\end{figure}

The cross-system statistics at the bottom of Table~\ref{tab:overall-quality} make the ceiling effect on FLORES-200 explicit. Under GEMBA-DA, the cross-system standard deviation on FLORES-200 zh-en is 2.29 and the score range is 7.87 points, while on HardMTBench zh-en the standard deviation rises to 5.24 and the range to 15.74 points. The same pattern holds on the en-zh direction, with the standard deviation moving from 2.04 to 4.54 and the range from 8.37 to 14.37 points. HardMTBench roughly doubles the cross-system spread on the GEMBA-DA scale. Figure~\ref{fig:bench-distribution} visualises the spread side by side.

The effect on xCOMET-XXL is more nuanced. The xCOMET standard deviation on FLORES-200 zh-en is 1.17, while on HardMTBench zh-en it rises to 3.78, a factor of about three. Yet the absolute level of xCOMET-XXL also drops sharply, from 96.16 to 63.52 on the same direction. This drop reflects a known property of xCOMET on out-of-distribution sentences and indicates that the absolute xCOMET-XXL value should not be read off the FLORES scale on HardMTBench, but its cross-system separation remains informative. In aggregate, HardMTBench is the benchmark among the three where both metrics show the widest spread across the 22 systems under test.

\subsection{General Benchmarks Do Not Fully Capture Hard-Domain Capability}
\label{sec:rank-reorder}

We test the hypothesis that general benchmarks already rank systems in the same order as a domain-heavy benchmark by computing Spearman rank correlations across the 22 systems. The FLORES-GEMBA average versus the HardMTBench-GEMBA average yields $\rho=0.881$, while the WMT25 en-zh GEMBA versus the HardMTBench en-zh GEMBA yields $\rho=0.778$. On xCOMET-XXL, the corresponding correlations are $\rho=0.910$ and $\rho=0.684$. While the correlations are positive, they are far from a perfect ranking, and the gap is especially visible in the middle band where the most informative comparisons for model selection happen.

\begin{figure}[!htbp]
\centering
\includegraphics[width=0.98\columnwidth]{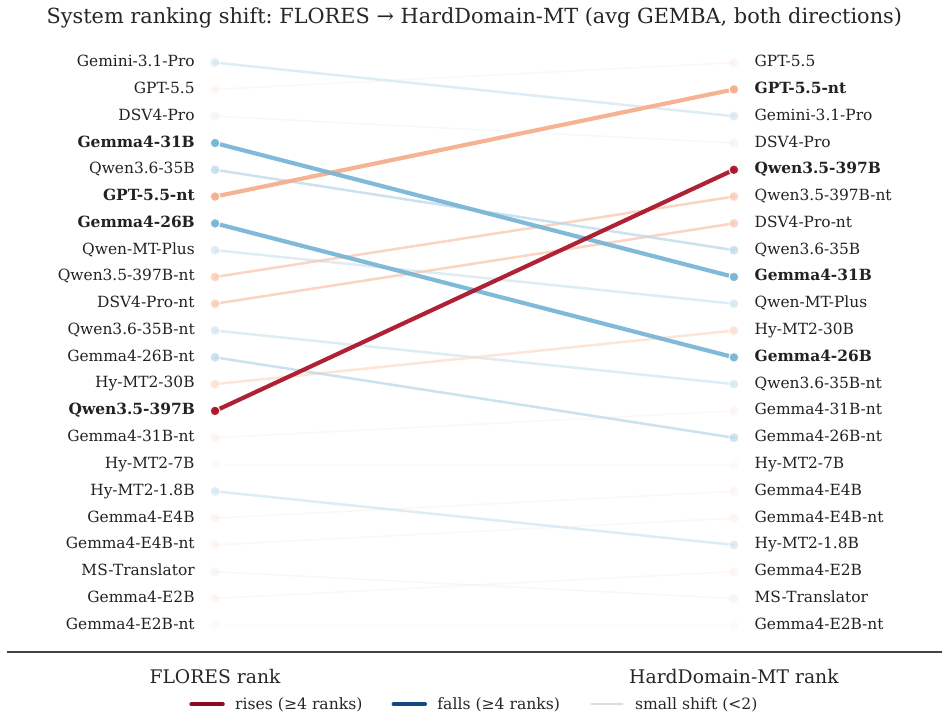}
\caption{System ranking shift from FLORES-200 to HardMTBench under GEMBA-DA averaged over zh-en and en-zh directions. Lines for systems with $\geq$4 rank positions of movement are drawn in bold and labelled in bold typeface, lines for shifts of $\geq$2 positions are drawn at intermediate strength, and the remaining lines are drawn in light grey.}
\label{fig:rank-slope}
\end{figure}

The largest rank movements are illustrative. Gemma4-31B ranks fourth on FLORES-GEMBA zh-en but sixth on HardMTBench-GEMBA zh-en. Microsoft Translator ranks 20th on FLORES en-zh but 22nd on HardMTBench en-zh, with an absolute gap of 12.7 GEMBA points opening up on the harder benchmark. Hy-MT2-30B-A3B drops from a strong FLORES profile (fourth on xCOMET zh-en) to a mid-table position on HardMTBench-GEMBA but climbs back to the top of HardMTBench-xCOMET, which indicates that the two metrics can reorder the same system on the same data. Figure~\ref{fig:rank-slope} traces all 22 systems between the two benchmarks. The most pronounced shifts ($\geq$4 ranks) are highlighted in saturated colour, while small reorderings remain visible but muted. Taken together, the rank reorderings suggest that performance on FLORES-200 or WMT25 alone is an incomplete proxy for capability on knowledge-intensive domain translation.

\subsection{Domain-Level Analysis}
\label{sec:domain}

Table~\ref{tab:domain-agg} reports cross-system averages per domain on HardMTBench, aggregated over zh-en and en-zh directions. GEMBA-DA and xCOMET-XXL are reported alongside terminology accuracy, which uses the per-sample terminology annotations described in Section~\ref{sec:pipeline}.

\begin{table}[t]
\centering
\scriptsize
\setlength{\tabcolsep}{7pt}
\renewcommand{\arraystretch}{0.95}
\begin{tabular}{lccc}
\toprule
Domain & GEMBA & xCOMET & Term.\,Acc.\ \\
\midrule
Finance & 92.04 & 75.46 & 58.45 \\
Law & 91.92 & 77.73 & 58.64 \\
Healthcare & 90.61 & 75.83 & 61.44 \\
Education & 90.31 & 70.74 & 49.50 \\
News & 89.25 & 66.49 & 56.37 \\
Military & 87.97 & 71.79 & 53.59 \\
Sci.\,\&\,Tech.\ & 87.67 & 72.65 & 66.07 \\
Sports & 87.43 & 67.38 & 55.50 \\
Books~/~Papers & 86.67 & 67.65 & 56.03 \\
Gaming & 85.31 & 59.29 & 45.97 \\
Media & 82.83 & 62.35 & 54.57 \\
History & 75.67 & 39.03 & 44.79 \\
\bottomrule
\end{tabular}
\caption{Cross-system averages per domain on HardMTBench, averaged over zh-en and en-zh directions. xCOMET-XXL is reported as a percentage, and terminology accuracy is the percentage of annotated terminology pairs whose target term appears in the translation after surface normalisation.}
\label{tab:domain-agg}
\end{table}

We highlight three observations. First, translation quality and terminology accuracy do not move together. Education has a high GEMBA score of 90.31 but a low terminology accuracy of 49.50, which indicates that translations can read smoothly while missing the controlled vocabulary that a domain practitioner expects. Second, the same gap appears in the opposite direction on Sci.\,\&\,Tech., where GEMBA is moderate (87.67) but terminology accuracy is the highest of all domains at 66.07, supported by stable standard naming for technical concepts. Third, History sits at the bottom on all three metrics, with GEMBA 75.67, xCOMET 39.03 and terminology accuracy 44.79. The combination of historical document language, person names, classical official titles and culturally loaded vocabulary remains the hardest region for current systems. Figure~\ref{fig:term-heatmap} (Section~\ref{sec:term-joint}) shows the analogous picture for terminology accuracy across systems and domains.

\subsection{Hardness as a Difficulty Proxy}
\label{sec:buckets-joint}

The per-sample hardness score $H$ from Equation~\ref{eq:hardness} lets us probe the relationship between difficulty and translation quality directly. We group the 22 systems into three tiers by their mean GEMBA-DA on HardMTBench (top tier of 7 systems including GPT-5.5 and Gemini 3.1 Pro, mid tier of 8 systems, and lower tier of 7 systems such as Hy-MT2-1.8B and Microsoft Translator) and partition the test set into equal-width hardness buckets of five points. From buckets 55--60 to 85--90, the top-tier mean GEMBA-DA falls by 5.4 points (91.4 to 86.0), the mid tier by 9.1 points (89.1 to 80.0), and the lower tier by 18.4 points (81.4 to 63.0). The widening gap with hardness suggests that hard samples carry the most discriminative signal across systems.

\begin{figure}[!htbp]
\centering
\includegraphics[width=\columnwidth]{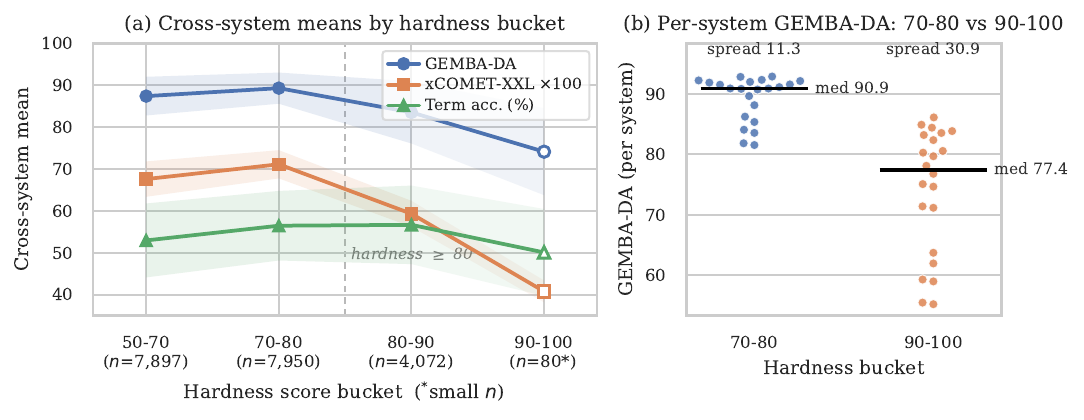}
\caption{Hardness bucket analysis. (a) Cross-system means of GEMBA-DA, xCOMET-XXL ($\times 100$) and term accuracy across four hardness buckets ($\pm$1 SD). The dashed line marks hardness $\geq 80$. Hollow markers at the 90--100 endpoint denote the small-sample bucket ($n=80$ per system) and use a lighter SD band. (b) Per-system GEMBA-DA for bucket 70--80 vs 90--100, with medians (black bars) and per-bin spread. The cross-system spread grows from 11.3 to 30.9 between the two buckets.}
\label{fig:hardness-bucket}
\end{figure}

Figure~\ref{fig:hardness-bucket} shows cross-system means across four hardness buckets (we merge the sparse 50--60 segment with 60--70 since it contains only 434 items per system and is not statistically informative on its own) together with the per-system distribution in the hardest bucket. The relationship is not strictly monotone in the easy half of the range: bucket 50--70 sits slightly below bucket 70--80 on GEMBA-DA, which reflects that the hardness score $H$ aggregates domain knowledge density, translation difficulty and terminology density rather than predicting quality directly. The pattern past hardness 80 is more pronounced. From bucket 70--80 to bucket 90--100, GEMBA-DA drops by 15.2 points (89.30 to 74.13), xCOMET-XXL by 30.4 points (71.12 to 40.77), and term accuracy by 6.4 points (56.45 to 50.07), and the cross-system spread on GEMBA-DA grows from 11.3 to 30.9. The 90--100 bucket carries an average of only 80 items per system, so its absolute values should be read as indicative. The across-bucket trend, however, is also supported by the much larger 80--90 bucket.

\subsection{Term-Level Diagnostic and Joint Reporting}
\label{sec:term-joint}

\begin{figure*}[t]
\centering
\includegraphics[width=0.66\textwidth]{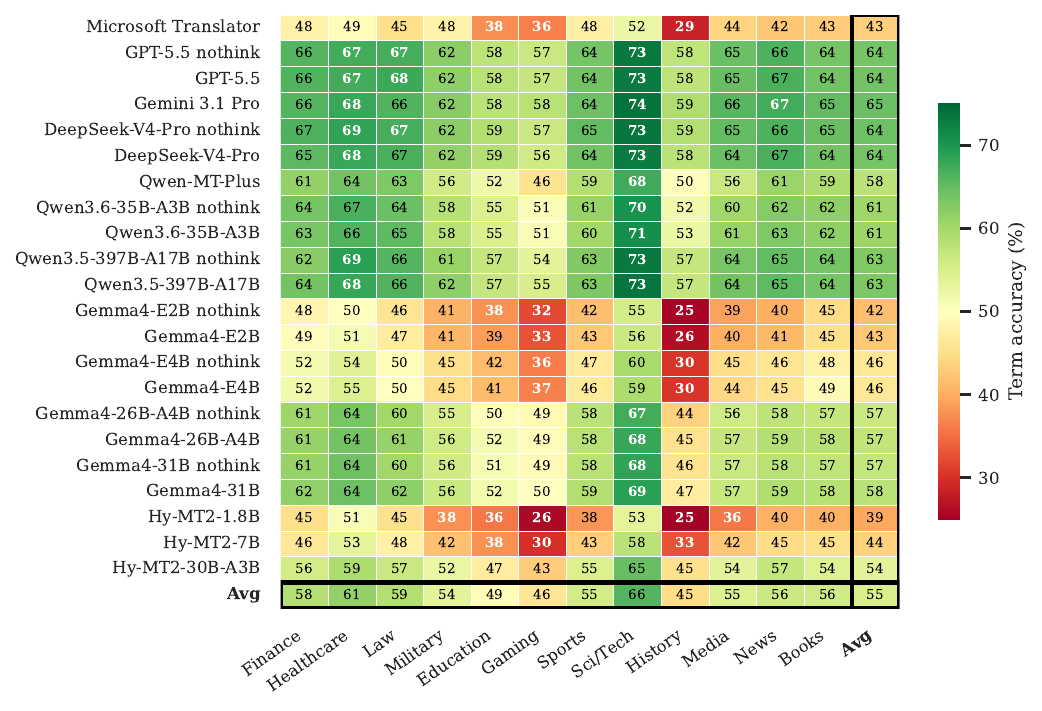}
\caption{Term accuracy (\%) on 22 systems $\times$ 12 HardMTBench domains, with marginal row/column means. Closed/API systems (GPT-5.5, Gemini 3.1 Pro, DeepSeek-V4-Pro) cluster around 64\%, while smaller open systems sit in the 40--46\% range. Sci.\&Tech.\ is easiest on terminology, History and Gaming are hardest.}
\label{fig:term-heatmap}
\end{figure*}

Quality scores compress a translation into a single number and tend to under-report failures that are localised to a few content-bearing tokens. We complement GEMBA-DA and xCOMET-XXL with a terminology-level diagnostic computed from the same span annotations that drive the hardness score. For each item, term accuracy is the fraction of source-side terminology spans whose target-side surface forms appear in the system output after light normalisation. This is a strict proxy that does not allow paraphrasing and should be read as a lower bound on terminology fidelity. Figure~\ref{fig:term-heatmap} shows term accuracy for all 22 systems across the 12 domains, with marginal averages on the right and at the bottom.

The term-level signal reorders the leaderboard noticeably. Gemini 3.1 Pro tops the term accuracy ranking at 64.52\% despite trailing GPT-5.5 on overall GEMBA-DA. The Hy-MT2 family illustrates the same point from below: Hy-MT2-30B-A3B reaches 89.15 on HardMTBench GEMBA-DA but only 53.62\% on term accuracy, and Hy-MT2-7B records the largest gap between strong xCOMET-XXL (71.4) and weak term accuracy (43.58\%) among all systems. The cross-system range on term accuracy is 25.0 points (39.50\%--64.52\%), which is larger than the 14--16 point GEMBA-DA range on the two HardMTBench directions and much larger than the 7--8 point range on FLORES-200. Terminology compliance and overall fluency-adequacy do not move together on hard-domain text.

Beyond terminology, GEMBA-DA and xCOMET-XXL themselves pick up partially complementary signals. Their Spearman correlation across the 22 systems on HardMTBench averaged over directions is $\rho=0.704$, which is meaningfully below the within-FLORES correlation. When all 44 (system, direction) cells are treated separately, the per-cell Spearman drops to $\rho=0.545$, and the en-zh cloud sits above the zh-en cloud on the xCOMET axis with mean xCOMET values of 0.709 and 0.635 respectively. This 7.4-point gap between directions does not appear on FLORES-200 to anywhere near the same extent and indicates a direction-dependent behaviour of xCOMET-XXL on out-of-distribution domain text. We therefore recommend that HardMTBench reports include three signals together: GEMBA-DA for fluency-adequacy, xCOMET-XXL for an independent reference-based view (interpreted alongside the direction), and term accuracy for terminology fidelity. Either of the first two metrics alone leaves a non-trivial fraction of the diagnostic signal unobserved.

\section{Related Work}
\label{sec:related}

\paragraph{General multilingual MT benchmarks.} FLORES-200~\citep{nllb2022flores}, NTREX-128~\citep{federmann2022ntrex} and the WMT general tasks~\citep{kocmi2023wmt23,kocmi2024findings} provide broad coverage with news and Wikipedia content. WMT24++~\citep{deutsch2025wmt24pp} expands to 55 languages and \citet{taguchi2025floresplusleftbehind} document remaining gaps for low-resource languages even after FLORES+ extensions. The per-system spread on the dominant pairs is now compressed. Top systems on FLORES Chinese-English cluster within 7--8 GEMBA points, which constrains the headroom for further diagnosis. HardMTBench targets the difficult tail of Chinese-English rather than language breadth.

\paragraph{Domain, terminology and robust MT benchmarks.} WMT Biomedical~\citep{neves2022wmt22biomedical}, TICO-19~\citep{anastasopoulos2020tico} and the WMT24 Discourse-Level Literary track~\citep{wang2024findingsdiscourselit} target individual domains, and recent work also covers multi-domain settings~\citep{hu2024multidomain_arxiv}, cultural and dialectal coverage~\citep{vandoren2026culturalnuance,yu2024wuchinesemt} and Chinese social-media text~\citep{zhao2026chinesesocialmt}. The WMT terminology task~\citep{alam2021wmt21term,dinu2019terminology} evaluates whether systems follow provided dictionaries and MTNT~\citep{michel2018mtnt} stresses noisy user-generated text. HardMTBench differs by spanning 12 difficulty-screened domains under a unified pipeline with balanced sub-domain quotas, evaluating natural translation of terminology-rich text without supplying term lists at inference while still annotating terminology so downstream metrics can be measured.

\paragraph{LLM-era MT evaluation and difficulty-aware testing.} LLM-as-judge protocols such as GEMBA-DA~\citep{kocmi2023gemba} and GEMBA-MQM~\citep{kocmi2023gembamqm} complement reference-based metrics, with the WMT24 metrics task~\citep{freitag2024llmjudge} and \citet{schmidt2026watchmen} documenting failure modes of automatic metrics on unseen and out-of-distribution domains, while \citet{issaka2026translationproxy} use translation as a scalable proxy for broader multilingual evaluation. \citet{kocmi2024findings} note that the LLM era compresses scores on standard sets while leaving failure modes intact. HardMTBench combines LLM-based scoring with rule-based features (terminology density, structural depth) and stratifies across 12 domains, and we report GEMBA-DA and xCOMET-XXL jointly because Section~\ref{sec:buckets-joint} shows they provide complementary signals on difficult text.

\section{Conclusion}
\label{sec:conclusion}

We presented HardMTBench, a difficulty-aware diagnostic benchmark for bidirectional Chinese-English domain translation, covering 12 knowledge-intensive domains with 10{,}000 hand-curated pairs (20{,}000 directional items), a transparent hardness score and terminology annotations. Across 22 systems on FLORES-200, WMT25 and HardMTBench, HardMTBench roughly doubles the cross-system GEMBA spread of FLORES-200, reorders system rankings, and surfaces domain-specific weaknesses in terminology compliance and historical-text translation. We hope HardMTBench complements rather than replaces FLORES-200 and WMT, and that the construction pipeline transfers to other language pairs and knowledge-intensive domains.

\section*{Limitations}

HardMTBench currently covers only Chinese-English, and the construction pipeline has not been validated on other language pairs. The 12 chosen domains reflect Chinese deployment scenarios and may underrepresent regions that are central in other markets. A subset of the released material is drawn from sources that are time-sensitive, for example news and audiovisual media, and the test set will need refresh cycles to remain representative. The hardness fusion rule in Equation~\ref{eq:hardness} is a transparent baseline rather than a learned ranker, and alternative weightings or learned hardness models could change the difficulty ordering. The two reported metrics, GEMBA-DA with gpt-oss-120b as the judge and xCOMET-XXL, inherit the biases of their underlying models, and human meta-evaluation of these metrics on hard domain text remains an open direction.

\section*{Ethics Statement}

The source material in HardMTBench is drawn from publicly available Chinese source text and is filtered to remove personal data, sensitive content and low-quality samples through the Stage 2 LLM judge and the human verification round described in Section~\ref{sec:verification}. Human annotators were trained bilingual speakers, were informed of the purpose of the work and received standard compensation. The release includes only aggregated system scores and the test set itself, and does not include individual model outputs that could enable re-identification of evaluated systems beyond their official release names.

\bibliography{custom}

\end{document}